\documentclass[letterpaper, 10 pt, conference]{ieeeconf}  % Comment this line out if you need a4paper

\IEEEoverridecommandlockouts                              % This command is only needed if 
\usepackage{color}
\usepackage{graphics} % for pdf, bitmapped graphics files
\usepackage{epsfig} % for postscript graphics files
\usepackage{mathptmx} % assumes new font selection scheme installed
\usepackage{times} % assumes new font selection scheme installed
\usepackage{amsmath} % assumes amsmath package installed
\usepackage{amssymb}  % assumes amsmath package installed
\usepackage{bm}
\usepackage{url}
\usepackage{balance}
\definecolor{xf}{RGB}{69,137,148}

\title{\LARGE \bf Fast Road Segmentation via Uncertainty-aware Symmetric Network}

\author{Yicong Chang$^{1,\dag}$, Feng Xue$^{1,\dag}$, Fei Sheng$^{1}$, Wenteng Liang$^{1}$, and Anlong Ming$^{1,*}$% <-this % stops a space
	\thanks{\dag Equal Contribution}
	\thanks{*Corresponding Author}
	\thanks{$^{1}$Beijing University of Posts and Telecommunications, Beijing, China,
		{\tt\small \{yicongchang,xuefeng,shengfei,liangwenteng,mal\}
		@bupt.edu.cn}}%
	\thanks{This work was supported by the national key R \& D program intergovernmental international science and technology innovation cooperation project 2021YFE0101600 and Excellent Ph.D. Students Foundation CX2020114.}
}

\begin{document}

\maketitle
\thispagestyle{empty}
\pagestyle{empty}

\begin{abstract}
The high performance of RGB-D based road segmentation methods contrasts with their rare application in commercial autonomous driving,
which is owing to two reasons:
1) the prior methods cannot achieve high inference speed and high accuracy in both ways;
2) the different properties of RGB and depth data are not well-exploited,
limiting the reliability of predicted road.
% Previous RGB-D based road segmentation methods have achieved great performance.
% However, there remain two critical issues:
% 1) existing methods are computationally intensive and can’t meet real-time requirements;
% 2) the different characteristics of RGB and depth data have not been well exploited.
In this paper,
based on the evidence theory,
an uncertainty-aware symmetric network (USNet) is proposed to achieve a trade-off between speed and accuracy by fully fusing RGB and depth data.
Firstly,
% Our network explicitly model uncertainty of prediction according to the evidence theory.
% In our network,
cross-modal feature fusion operations,
which are indispensable in the prior RGB-D based methods,
are abandoned.
We instead separately adopt two light-weight subnetworks to learn road representations from RGB and depth inputs.
The light-weight structure guarantees the real-time inference of our method.
Moreover,
a multi-scale evidence collection (MEC) module is designed to collect evidence in multiple scales for each modality,
which provides sufficient evidence for pixel class determination.
Finally,
in uncertainty-aware fusion (UAF) module,
the uncertainty of each modality is perceived to guide the fusion of the two subnetworks.
% and segmentation result is generated by an uncertainty-aware fusion module.
Experimental results demonstrate that our method achieves a state-of-the-art accuracy with real-time inference speed of \emph{43+} FPS. The source code is available at \url{https://github.com/morancyc/USNet}.

% \textcolor{red}{
% Existing road segmentation methods based on RGB-D images have achieved great performance. However, there is a problem that has not been well studied: both RGB and depth sources contain noisy features, and the existing RGB-D based methods do not eliminate the noisy features of the two modalities, which may leads to false detection. In this work, we propose a new road segmentation network to tackle this problem. We adopt a two-stream CNN which separately make predictions based on RGB and depth modalities. Furthermore, we use the multi-scale prediction strategy to reduce influence of noise in each modality. Finally, the detection results of two modalities are fused by a evidential opinions fusion module. By this means, we explicitly model the uncertainty of the prediction result, so as to eliminate the influence of noise in a certain modality. Our network has achieved competitive performance on the KITTI benchmark, and experiments on the Cityscapes dataset proved the generalization of our method.}

\end{abstract}

\section{INTRODUCTION}
% Road segmentation is a significant component in self-driving system.
% It aims to classify each pixel of an image as drivable or undrivable.
% Thanks to the blossom of deep learning,
% many works based on deep neural network (DNN) have gained high accuracy on road segmentation.
% According to the sensors used,
% i.e.,
% 3D LiDAR,
% monocular and stereo camera,
% they are divided into three groups:
% RGB based methods,
% RGB-LiDAR based methods,
% and RGB-D based methods.
% The first group \cite{Efficient2016, RBNet, RBANet, Embedding_contour} \xf{only demands monocular RGB camera,}
% but the lack of geometric information limits their reliability in real applications.
% \xf{To boost reliability,}
% RGB-LiDAR based methods \cite{LidCamNet, PLARD} employ monocular camera simultaneously with 3D LiDAR that captures scene depth,
% but they are hard to be widely applied due to the high cost of 3D LiDAR.
% For a better trade-off between reliability and cost,
% RGB-D based methods \cite{SNE-RoadSeg, NIM-RTFNet, DFM-RTFNet, SNE-RoadSeg+} utilize stereo camera to obtain RGB and depth data.
% \xf{They are valued by the community as the low cost and high data resolution.}

% Road segmentation aims to classify each pixel of an image as drivable or undrivable,
% and provides the drivable region of vehicle to other modules in self-driving system for supporting safe navigation.
Road segmentation aims to classify each pixel of an image as drivable or undrivable and provides the drivable region to other modules in self-driving system for safe navigation.
% \xf{It is vital in applications of self-driving and
% mobile robot \cite{Objectness_aware_tracking, YuZhou-IJCV2016-SFVT, NIPS2012_3e313b9b}}
% Thus, It is a fundamental component in self-driving system and advanced driver assistant system,
Obviously,
both high efficiency and reliability are necessary for road segmentation.
In recent years,
many algorithms have been proposed to segment road based on vision sensors,
i.e.,
3D LiDAR, monocular and stereo camera.
Compared to the works using low-cost monocular \cite{Efficient2016, MultiNet, RBANet} and expensive LiDAR sensors \cite{LidCamNet, CLCFNet, PLARD},
the road segmentation methods using affordable stereo camera \cite{SNE-RoadSeg, NIM-RTFNet, SNE-RoadSeg+},
namely RGB-D based methods,
achieve a better trade-off between cost and reliability. Thus, they are valued by the community recently.

\begin{figure}
    \setlength{\abovecaptionskip}{-0.05cm}
    \centering
    \includegraphics[width=1\linewidth]{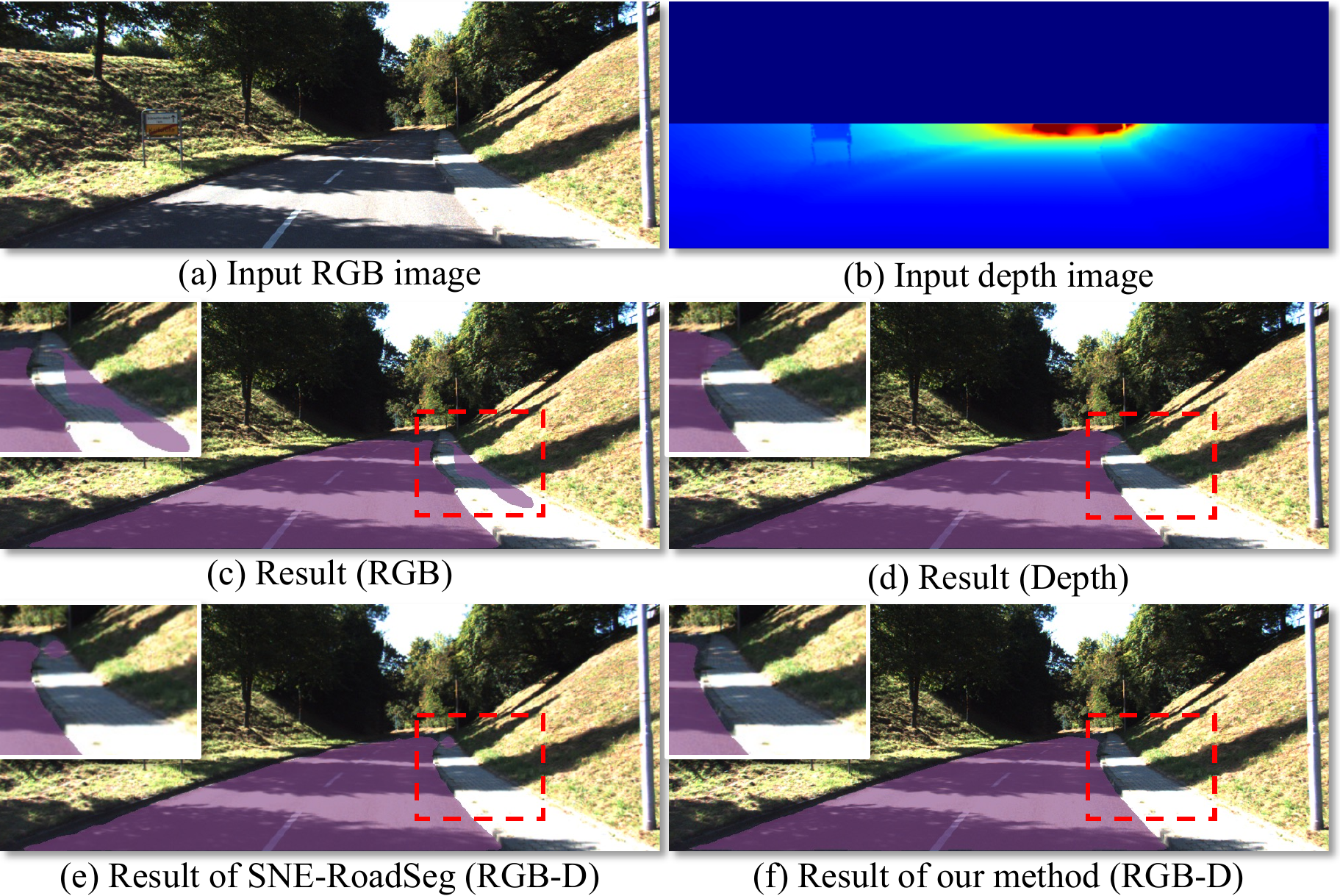}
    \vspace{-0.4cm}
    \caption{Segmentation results of different models.
    (a) RGB image.
    (b) Depth image.
    (c) Result of the model using RGB as input.
    (d) Result of the model using Depth as input.
    (e) Result of SNE-RoadSeg \cite{SNE-RoadSeg}.
    (f) Result of our method.
    The segmented road are marked in purple,
    and the images are zoomed to clearly show the details.}
    \vspace{-0.4cm}
    \label{fig:intro}
\end{figure}

However,
the prevalence of RGB-D based methods contrasts with their rare application.
One of the reasons is that although a great breakthrough in performance has been made,
the previous RGB-D based road segmentation methods still cannot run in real-time.
% the previous RGB-D based road segmentation methods still cannot achieve high accuracy while meeting real-time computation.
Another reason is that the previous methods are limited on the reliability of predicted road,
as they cannot fully utilize the particular characteristics of RGB and depth data.
% \xf{As two basic modalities expressing the 2D space and 3D space,
% RGB and depth data show different characteristics in street scenario.
% RGB data is sensitive to light-dark contrast,
% e.g., over-exposure and shadow region on the street,
% but depth data is sensitive to distance change,
% e.g. road boundary.
% It is hard for convolution neural networks (CNNs) to parse their characteristics.}
Theoretically,
RGB and depth data are two basic visual modalities that express the 2D space and 3D space, respectively.
They hold different characteristics:
\emph{RGB data is sensitive to light-dark contrast in the 2D spatial domain,
% e.g., over-exposure and shadow region on the road,
while depth data is sensitive to distance change in the 3D spatial domain.}
% e.g. road boundary.
How to perceive the characteristic difference is a key issue for RGB-D based road segmentation.
% To fuse the two modalities in inference,
The previous methods \cite{SNE-RoadSeg, NIM-RTFNet, SNE-RoadSeg+} commonly merge RGB and depth data by summing features of the two modalities to form a unified representation of road.
However,
RGB and depth features are indiscriminately fused in this design,
bringing in conflict of feature representation in the sensitive region of RGB and depth data.
% introducing wrong feature representations of these two modalities for the road,
It would lead to wrong segmentation inside these sensitive regions.
To better explain this issue,
we trained two networks that utilize RGB and depth image as input respectively.
As shown in Fig.\ref{fig:intro} (c) (d),
the RGB-based and depth-based models show different false positives,
which are caused by their different representations of the road.
This conflict cannot be eliminated by adding the two modality features \cite{SNE-RoadSeg},
as shown in Fig.\ref{fig:intro} (e).

% the prediction using RGB data have false positives nearby the high-contrast region,
% and the prediction using depth data suffers from false positives near the blurred road boundary.
% As shown in Fig.\ref{fig:intro} (e),
% it cannot eliminate the wrong prediction by adding the two modal features.

To address this issue,
we abandon the mainstream idea of boosting the discriminative power of features adopted by many works \cite{Video_Text_Tracking, YuZhou-IJCV2016-SFVT, NIPS2012_3e313b9b, Jianxiang-ICCV2017-OLP, Objectness_aware_tracking} on other vision tasks.
% 2014ONLINE
In this paper,
% the motivation is the conjecture that models generally obtain a high uncertainty for the wrong predicted area.
we introduce evidence theory to fuse RGB and depth data 
% under the guidance of uncertainty
and gain state-of-the-art performance with low computational cost.
% \cyc{
% Previous works on various vision tasks \cite{Video_Text_Tracking, ZheLiu-AAAI2019-TANet, 2014ONLINE} commonly focus on extracting discriminative features from image to eliminate the interference of noise.
To this end,
an uncertainty-aware symmetric network (USNet) is proposed,
which perceives the uncertainties of each modality,
and incorporates the uncertainties into the fusion process.
% While we propose an uncertainty-aware symmetric network (USNet),
% which perceives the uncertainties of RGB and depth images,
% and incorporates the uncertainties into the fusion process.}
Firstly,
the cross-modal feature fusion utilized by previous works \cite{SNE-RoadSeg,NIM-RTFNet,SNE-RoadSeg+} is abandoned.
% We instead design two subnetworks with a same light-weight structure to learn road representations from RGB and depth inputs,
% avoiding feature-level conflict.
We instead design two light-weight subnetworks to learn road representations from RGB and depth inputs,
avoiding feature-level conflict.
Secondly,
for each modality,
a multi-scale evidence collection (MEC) module is proposed to collect evidence of a pixel belonging to road or non-road in multiple scales.
% a multi-scale evidence collection module (MSC) is proposed to segment road in multiple scales,
% and combine all results to obtain a low-uncertainty road area as completely as possible.
Thirdly,
based on the evidence given by the two subnetworks,
an uncertainty-aware fusion (UAF) module is designed to perceive the uncertainty of each modality,
and determine the probability and uncertainty that a pixel belongs to road or non-road.
In practical application,
the perceived uncertainty can be provided to the self-driving system for further obstacle discovery \cite{TOD_2019,TOD_2020}.
% an uncertainty-aware fusion (UAF) module is designed to fuse the outputs of the two networks based on uncertainty.
% For each subnetwork,
% it uses a triplet to represent a pixel,
% i.e., two of this triplet showing the degree of the pixel belonging to the non-road or the road,
% and one for the uncertainty of other two values.
% Following Dempster’s Combination Rule \cite{Evidential_deep_learning},
% we assign a higher weight to the triplet of the two modalities with lower uncertainty,
% and perform a weighted fusion of the two modal result of the road.
As shown in Fig. \ref{fig:intro} (f),
by considering the uncertainty of each modality,
our method eliminates the wrong segmentation completely.
Our network achieves comparable performance with the state-of-the-art methods with an inference speed of \emph{43}$+$ FPS,
\emph{4}$\times$$\sim$\emph{10}$\times$ faster than the former state-of-the-art methods.
The contributions of our approach lie in:

\begin{itemize}
\item
The conflict of RGB and depth data in feature space is revealed,
which guides us to propose a light-weight symmetric paradigm for RGB-D based road segmentation to avoid the feature conflict of different modalities.

% A light-weight symmetric paradigm is proposed for RGB-D road segmentation,
% which avoids the feature conflict of different modalities.
% A multi-scale evidence collection module is proposed to obtain road area as reliable as possible by combining multi-scale segmentation.

\item
% An uncertainty-aware fusion module is proposed to eliminate the wrong prediction in fusion by employing the low-uncertainty result.
% Based on the evidence theory,
% a multi-scale evidence collection (MEC) module is proposed to obtain sufficient evidence of road and non-road in multiple scales,
% so that the well-designed uncertainty-aware fusion (UAF) module makes full use of the characteristics of RGB and depth data.
Based on the evidence theory,
a multi-scale evidence collection (MEC) module is proposed to obtain sufficient evidence of road and non-road,
so that the well-designed uncertainty-aware fusion (UAF) module makes full use of the characteristics of RGB and depth data.

\item
Our method gains state-of-the-art accuracy on both the KITTI benchmark \cite{KITTI} and the Cityscapes dataset \cite{Cityscapes},
while with a low cost of computation and parameter.

% Our method gains state-of-the-art accuracy on both the KITTI benchmark \cite{KITTI} and the Cityscapes dataset \cite{Cityscapes},
% while with a low cost of computation and parameter.

\end{itemize}

\section{Related Work}

\subsection{Road Segmentation}

Road segmentation has been studied for decades.
It is vital in self-driving and
mobile robot applications \cite{RoadTrack,2014ONLINE}.
% The introduction of deep learning has boosted the development of this task.
After Long \emph{et al.} \cite{FCN} proposed the pioneering work,
i.e., Fully Convolutional Network (FCN),
for semantic segmentation,
many methods based on the FCN framework have arisen.
% \cite{Efficient2016,RBNet,RBANet,Embedding_contour,LidCamNet,CLCFNet,PLARD,SNE-RoadSeg,NIM-RTFNet,DFM-RTFNet,SNE-RoadSeg+}.
Previous methods can be divided into three groups:
RGB-based methods \cite{RBNet, Embedding_contour},
RGB-LiDAR based methods \cite{LidCamNet,CLCFNet},
and RGB-D based methods \cite{SNE-RoadSeg,SNE-RoadSeg+,DFM-RTFNet}.

\begin{figure*}
    \centering
    \vspace{0.15cm} %调整图片与上文的垂直距离
    \includegraphics[width=0.954\linewidth]{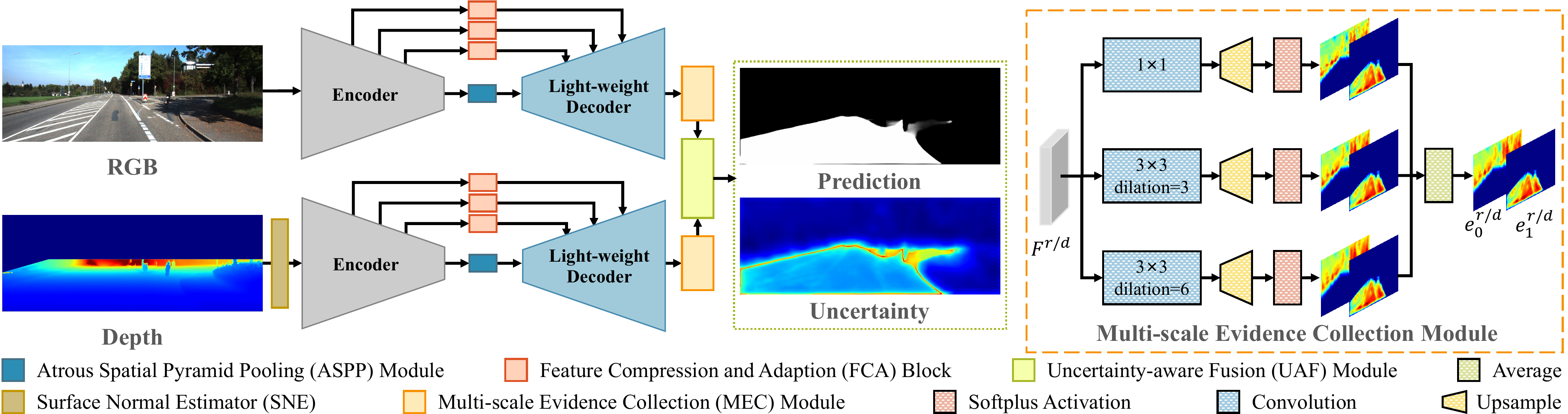}
    % \vspace{-15pt}
    \caption{Network architecture of USNet.
    It uses SNE \cite{SNE-RoadSeg} to calculate the surface normal.
    Our network is composed of two subnetworks that take RGB and depth image as input, respectively.
    The detailed structure of the multi-scale evidence collection (MEC) module is shown in the right.}
    \vspace{-0.25cm}
    \label{fig:network}
\end{figure*}

The RGB-based methods \cite{MultiNet, RBANet} improve the accuracy of road segmentation by introducing extra cue,
such as boundary \cite{hed,Jianxiang-ICCV2017-OLP} and object \cite{fasterrcnn,ZheLiu-AAAI2019-TANet}.
However,
they may fail when there exist illumination variations in the RGB image.
Considering the above problem,
many approaches use LiDAR or Depth data as a supplement \cite{LidCamNet,CLCFNet,DFM-RTFNet}.
% As for RGB-LiDAR based methods, Caltagirone \emph{et al.} \cite{LidCamNet} designed a cross fusion FCN to  integrate information from the LIDAR and RGB inputs. Chen \emph{et al.} \cite{PLARD} first transformed the LiDAR data to the perspective view and then adapted LiDAR features to visual features through a cascaded fusion structure.
As for the RGB-D based methods,
Wang \emph{et al.} \cite{DFM-RTFNet} propose an effective and efficient data-fusion strategy named DFM to fuse RGB and depth features.
Based on the hypothesis that the road area in one image is coplanar,
\cite{SNE-RoadSeg,NIM-RTFNet,SNE-RoadSeg+} propose to estimate surface normal from the depth image and utilize the surface normal as the input of the network.
However,
the different characteristics of RGB and depth data have not been well utilized by the above-mentioned methods, thus leaving room for performance improvement.
% In contrast, our approach explicitly models the uncertainty of RGB and depth and combines the two predictions in an uncertainty-aware way. Thus the complementary of RGB and depth can be well exploited.

\subsection{Evidence-based Learning}

Vanilla neural networks make predictions by a deterministic learning pipeline.
However,
accurate predictions achieved on datasets are not sufficient to cope with the real world.
It is also significant to obtain the reliability of the prediction.
% a good learner not only needs to make accurate prediction,
% but also needs to know the reliability of the prediction.
To this end,
evidence-based methods propose a way to model the uncertainty of prediction \cite{Evidential_deep_learning,Trusted_fusion}.
The idea of evidence-based learning comes from Dempster–Shafer Theory of Evidence (DST) \cite{DST}.
Subjective Logic (SL) \cite{Subjective_Logic} defines a framework that formalizes DST’s notion of belief assignments as a Dirichlet Distribution.
Based on the SL,
Sensoy \emph{et al.} \cite{Evidential_deep_learning} propose a framework to model the uncertainty of classification tasks.
After that,
Han \emph{et al.} \cite{Trusted_fusion} accomplish trusted classification by using Dempster’s combination rule to fuse classifications of multiple views.
% To the best of our knowledge,
% evidence based learning has not been exploited in road segmentation,
% and we are the first to employ evidence based fusion strategy to realize uncertainty-aware RGB-D road segmentation.
Inspired by \cite{Trusted_fusion},
we follow the evidence-based fusion strategy to realize uncertainty-aware RGB-D road segmentation.

\section{Background}

% Since our method is built according to the Dempster-Shafer Theory of Evidence (DST) \cite{DST},
% we give a brief review to DST in this section.

% DST directly models uncertainty of classification tasks by assigning \prob{belief mass to any possible class} and an overall uncertainty to the whole frame.
% Subjective Logic (SL) \cite{Subjective_Logic} defines a framework that formalizes DST’s notion of belief assignments as a Dirichlet Distribution.
% \prob{In this framework,
% the output of the network is treated as evidences collected for each possible class and are associated with the concentration parameters of the Dirichlet distribution.}

The theoretical basis of uncertainty modeling in our method is Subjective Logic (SL) \cite{Subjective_Logic}.
% which defines a framework to obtain uncertainty of a prediction.
In this section, we give a brief review of it.

For $K$-class classification,
a front-end model is utilized to extract evidence,
namely,
information that supports a sample to be classified into a certain class.
Based on the extracted evidence,
% given the evidences extracted by a front-end model,
SL assigns a probability (called belief mass) to each class,
and generates an overall uncertainty (called uncertainty mass) of this assignment.
% based on evidence collected from data.
The belief assignment is formalized as a Dirichlet distribution in SL, and the concentration parameters of the Dirichlet distribution are related to the evidence.
% The evidence in this scope denotes support information that in favor of a sample to be classified into a certain class and is related to the concentration parameters of Dirichlet distribution.
To be specific,
% for the $K$-class classification,
the evidence of each class is denoted as $e_{k}, \ k=1,...,K$,
where $e_{k} \ge 0$.
% we will obtain $e_{k}, \ k=1,...,K$, where $e_{k}\ge 0$ and it indicates the evidence collected for each class.
And the concentration parameters $\alpha_{k}$ of this Dirichlet distribution is pre-defined as $\alpha_{k}=e_{k}+1$.
Then the belief mass $b_{k}$ and the uncertainty $u$ can be computed by:
%inversely proportional to the total evidence
\begin{equation} \label{eq_1}
b_{k} =\frac{e_{k} }{S} =\frac{\alpha_{k}-1}{S} \quad  and  \quad u =\frac{K}{S}    
\end{equation}
where $S= {\sum_{k=1}^{K}}(e_{k} + 1)= {\sum_{k=1}^{K}} \alpha _{k}$ is the Dirichlet strength.
Intuitively,
the uncertainty $u$ is inversely proportional to the total evidence $\sum_{k=1}^{K}e_{k}$,
which means the more evidence collected,
the lower the uncertainty.
According to Eq. \ref{eq_1},
the $K$ belief masses $b_{k}$ and $u$ are all non-negative and their sum is one:
$u + \sum\nolimits_{k=1}^{K} b_{k} = 1$.
% The belief mass assignment can be interpreted as a subjective opinion.
Given the belief assignment,
the expected probability for the $k$-th class is the mean of the corresponding Dirichlet distribution:
\begin{equation} \label{eq_3}
\hat{p}_{k} = \frac{\alpha_{k}}{S} 
\end{equation}
% In practise,
% a CNN is capable of collecting evidence $e_{k}$ from an input sample,
% and the parameters $\alpha_{k}$ of the Dirichlet distribution are $\alpha_{k}=e_{k}+1$, thus we can calculate the class probabilities by Eq. \ref{eq_3}.
In this paper,
we use a CNN to collect evidence from input images,
and calculate the uncertainty of each pixel to guide the fusion of predicted road from RGB and depth data.

\section{METHODOLOGY}

In this section,
we propose an uncertainty-aware symmetric network (USNet),
which is illustrated in Fig. \ref{fig:network}.
Note that,
the depth image is processed by the surface normal estimator (SNE) \cite{SNE-RoadSeg} to generate a three-channel surface normal that is taken as the input for the depth subnetwork.
This transformation is important for our method as it provides a more effective representation of the road.
Different from the previous road segmentation methods \cite{SNE-RoadSeg, NIM-RTFNet,SNE-RoadSeg+} which conduct cross-modal feature fusion,
the USNet learns two feature representations of road from RGB and depth image by two independent subnetworks (see Sec. \ref{sec:OneNetwork}), respectively.
Then, sufficient evidence is collected from the well-designed multi-scale evidence collection (MEC) module (see Sec. \ref{sec:mse}).
% The unimodal subnetwork is illustrated in Sec. \ref{sec:OneNetwork}.
Finally, an uncertainty-aware fusion (UAF) module is introduced to determine the pixel class based on evidence of the two modalities
% fuse the segmentation results of the two modalities
(see Sec. \ref{sec:fusion}).
% \subsection{Network Overview} 
% In this work,
% we propose a symmetrical two-stream network to separately exploit RGB and depth image for road segmentation.
% For specific,
% the depth image is first processed by surface normal estimator (SNE) \cite{SNE-RoadSeg} to generate a three channel surface normal,
% and then it can be directly fed into a pre-trained encoder.
% The structure of USNet is illustrated in Fig. \ref{fig:network}.
% Different from previous road segmentation methods \cite{SNE-RoadSeg, NIM-RTFNet, DFM-RTFNet, SNE-RoadSeg+},
% the feature-level fusion between RGB and depth modalities is not incorporated in our network architecture.
% In contrast,
% our network consists of two independent streams and each stream adopts a U-Net-like \cite{UNet} structure to progressively aggregate multi-scale features from one modality for road segmentation.
% Moreover,
% we utilize a multi-scale prediction module at the end of each stream to reduce false detection caused by noise in each modality.
% At last,
% we design an evidential opinion fusion module to further fuse the detection results of the two modalities and to generate final prediction.

\begin{figure*}
    \setlength{\abovecaptionskip}{-0.1cm}
    \centering
    \vspace{0.15cm} %调整图片与上文的垂直距离
    \includegraphics[width=1\linewidth]{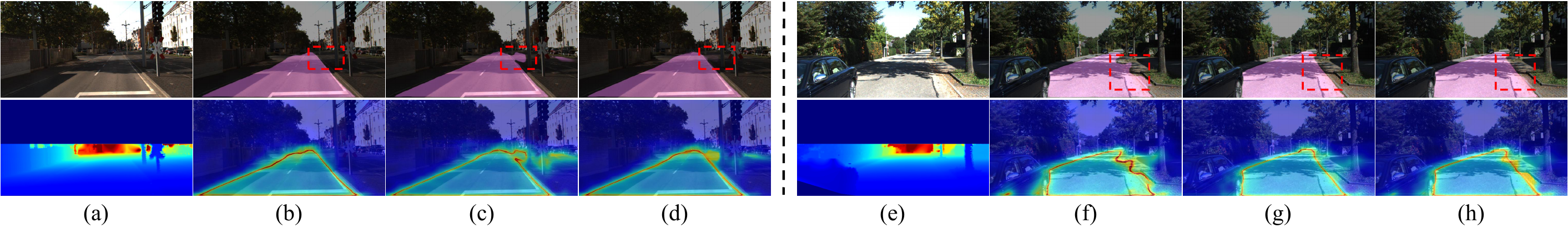}
    \caption{Visualization of uncertainty-aware fusion.
    (a) and (e): RGB and depth images.
    (b) and (f): $b_{1}^{\mathbf{r}}$ and $u^{\mathbf{r}}$ of the RGB subnetwork.
    (c) and (g): $b_{1}^{\mathbf{d}}$ and $u^{\mathbf{d}}$ of the depth subnetwork.
    (d) and (h): $b_{1}$ and $u$ after fusion.
    The density of the color map indicates the magnitude of value.
    }
    \vspace{-0.25cm}
    \label{fig:uncertainty}
\end{figure*}

\subsection{Unimodal Subnetwork to Obtain Road Representation}

\label{sec:OneNetwork}
In USNet,
the subnetwork for unimodal aims to extract the feature representation of road from an RGB or depth image.
% based on single modality input.
The subnetwork is mainly composed of an encoder and a light-weight decoder.
To be specific,
the ResNet-18 \cite{ResNet} is exploited as the encoder for low latency.
At the end of the encoder,
inspired by \cite{OFNet},
an atrous spatial pyramid pooling (ASPP) module \cite{Deeplab} is employed to perceive multi-scale context for road segmentation.
% The dilated rates of dilated convolutions are 6, 12 and 18, respectively.
Subsequently,
the side-outputs of the encoder are extended by several simple blocks,
called feature compression and adaptation (FCA) blocks.
In an FCA block,
a $1\times1$ convolution is applied to reduce the number of channels to 64,
followed by a channel-wise attention extractor \cite{SENet} to enhance the discriminative channels for road representation.
The FCA block helps the subnetwork to restore scene details by incorporating the compressed feature into the decoder.
% \prob{A unimodal road segmentation stream} aims to predict \prob{a confidence map of road} based on \prob{a single modal input}.
% For the encoder, we exploit ResNet-18 \cite{ResNet} as our backbone.
% At the end of encoder,
% an atrous spatial pyramid pooling (ASPP) module \cite{Deeplab} is \prob{incorporated}.
% The ASPP module consists of one $1\times1$ convolution layer and three dilated convolution layers with kernel size of $3\times3$ and dilation rate of 6, 12 and 18 respectively.
% \prob{With this module,
% our network is able to perceive the scene cue of different scales.}
% to achieve a fast inference speed,
% \prob{Then, the decoder part of the subnetwork follows a zero-parameter structure to achieve a fast inference speed.}
The decoder part of the subnetwork follows a zero-parameter structure to reduce model complexity.
It consists of three upsampling-summing stages to gradually upsample the feature to $1/4$ size of the original image.
Each stage straightforwardly sums the upsampled feature from the previous stage with the feature from the corresponding FCA block.
Note that the first stage takes ASPP output as the input.
And the last stage of the decoder yields the finest 64-channel feature.
In summary,
the light-weight design ensures the high inference speed of USNet.
% And the output size of the decoder is $1/4$ of the original image.

% The decoder gradually upsample the feature to the size of original image,
% and skip connections from high resolution feature maps in the encoder helps restore scene details.
% The skip connections in our network are implemented by the feature compression and adaptation (FCA) modules.
% A FCA module consists of a convolution layer and a channel-wise attention module. Specifically,
% given a feature map from the encoder,
% we first apply a $1\times1$ convolution to reduce the channels.
% Then,
% a channel-wise attention module is employed to extract correlations between channels and the adapted feature map is added to corresponding layer in the decoder. 
% To achieve a fast inference speed,
% the feature maps in the decoder are all compressed to 64 channels.

\subsection{Multi-scale Evidence Collection (MEC) Module}
\label{sec:mse}
To sufficiently gain evidence for judging a pixel's class,
we append a multi-scale evidence collection (MEC) module after each subnetwork.
The detailed structure of MEC is shown in Fig. \ref{fig:network}.
% We take the RGB stream as an example to elaborate the operations of the MSC.
% As shown in Fig. \ref{fig:network}, 
For the MEC of RGB subnetwork,
given the RGB feature output by the decoder, denoted as $F^{\mathbf{r}}$,
the MEC outputs two evidence maps $\{e_{0}^{\mathbf{r}}, e_{1}^{\mathbf{r}}\}$,
where $e_{0}^{\mathbf{r}}$ indicates the pixel-wise evidence of non-road,
and $e_{1}^{\mathbf{r}}$ for that of road.

Specifically,
MEC consists of three parallel paths.
In each path,
a convolution is first employed to gain a two-channel feature map at multiple scales.
The kernel sizes of the convolution in these paths are $1\times1$, $3\times3$, and $3\times3$,
and the dilation rates of the two $3\times3$ convolutions are $3$, $6$, respectively.
% The two channel of output feature is taken as evidence maps of road and non-road.
Then, an upsample layer is employed to upscale the two-channel feature map to the size of the original image.
% i.e., $H\times W$.
At the end of each path,
a softplus activation layer is used to realize the non-negativity of all feature values,
obtaining evidence maps $\{[e_{0}^{\mathbf{r}}]^h, [e_{1}^{\mathbf{r}}]^{h}\}$,
where $h=1,2,3$ is the index of the path.
% Each path can be formulated as:
% \begin{equation} \label{eq_4}
% \{[e_{0}^{\mathbf{r}}]^h, [e_{1}^{\mathbf{r}}]^{h}\} = \epsilon(up(conv_h(F^{\mathbf{r}})))
% \end{equation}
% where $conv$, $up$, and $\epsilon$ represent the convolution,
% upsample and softplus layer, respectively.
Then the final evidence of the RGB subnetwork is defined as the mean of these evidence maps in all paths:
\begin{equation} \label{eq_4}
e^{\mathbf{r}}_{k} = \frac{1}{3}\sum\nolimits_{h=1}^3 [e_{k}^{\mathbf{r}}]^h , \quad k=0,1
\end{equation}
For the depth subnetwork,
we use another MEC to capture evidence maps $\{e_{0}^{\mathbf{d}},e_{1}^{\mathbf{d}}\}$ from the depth feature.

By using the convolutions with different receptive fields,
MEC is able to extract multi-scale evidence,
which makes our method more reliable in determining each pixel's class.

\subsection{Uncertainty-aware Fusion (UAF) Module}
\label{sec:fusion}
% In previous road segmentation works, the last layer of the network commonly outputs a two-channel feature map and the final confidence map is achieved by using a softmax function on it. However, the output of softmax usually leads to over-confidence. In other words, the values in the final predicted confidence map are all close to 0 or 1 even for those false predictions. We would like the network to generate more reliable results. That is, for the hard areas in the image, the predicted results are expected to be close to $0.5$.

To accurately segment the road by considering modal characteristics,
an uncertainty-aware fusion (UAF) module is proposed,
which combines the evidence of RGB and depth modal under the guidance of uncertainty.
Specifically,
given the output evidence of MEC modules,
i.e., $\{e_{0}^{\mathbf{r}}, e_{1}^{\mathbf{r}}\}$ for RGB modality and $\{e_{0}^{\mathbf{d}}, e_{1}^{\mathbf{d}}\}$ for depth modality,
the probability of each pixel belonging to road is predicted by three steps.
% uncertainty is perceived and is incorporated into the fusion process by three steps.

The first step generates the belief masses and uncertainty of each modality. As illustrated in Sec. III, the belief assignment is formulated as a Dirichlet distribution.
% For the RGB subnetwork, giving evidence maps $\{e_{0}^{\mathbf{r}}, e_{1}^{\mathbf{r}}\}$,
For the evidence maps $\{e_{0}^{\mathbf{r}}, e_{1}^{\mathbf{r}}\}$ of RGB modality,
% \prob{the Dirichlet strength is formulated as $S^{\mathbf{r}} = e^{\mathbf{r}}_0+e^{\mathbf{r}}_1+2$,}
a Dirichlet strength map is defined as
% the Dirichlet strength is
$S^{\mathbf{r}}= {\sum_{k=0}^{1}}(e_{k}^{\mathbf{r}} + 1) = e^{\mathbf{r}}_0+e^{\mathbf{r}}_1+2$.
Then the belief masses $b_{k}^{\mathbf{r}}$ of non-road and road and the uncertainty map $u^{\mathbf{r}}$ are obtained by using Eq. \ref{eq_1}:
$b_{0}^{\mathbf{r}}=e_{0}^{\mathbf{r}}/S^{\mathbf{r}}$, $b_{1}^{\mathbf{r}}=e_{1}^{\mathbf{r}}/S^{\mathbf{r}}$,
and $u^{\mathbf{r}}=2/S^{\mathbf{r}}$.
Note that the '$/$' denotes pixel-wise division in this subsection.
We use $\{b_{0}^{\mathbf{r}}, b_{1}^{\mathbf{r}}, u^{\mathbf{r}}\}$ to denote the belief assignment of RGB subnetwork.
% We use $M^{\mathbf{r}}=\{b_{0}^{\mathbf{r}}, b_{1}^{\mathbf{r}}, u^{\mathbf{r}}\}$ to denote the belief mass assignment of RGB modalites.
For a pixel,
if its evidence of non-road and road, i.e., $e_{0}^{\mathbf{r}},e_{1}^{\mathbf{r}}$, are both small,
it means that we lack evidence of the RGB modality to determine the category of the pixel.
In this case, the pixel's uncertainty would be very large.
% Note that it is impossible for $e_{0}^{\mathbf{r}}$ and $e_{1}^{\mathbf{r}}$ to obtain a large value at the same time,
% the reason is illustrated in Sec. \ref{sec:loss}.
In the same way,
the belief assignment of depth subnetwork can be obtained: $\{b_{0}^{\mathbf{d}}, b_{1}^{\mathbf{d}}, u^{\mathbf{d}}\}$.

Referring to Dempster’s combination rule \cite{DST},
the second step merges the belief masses of two modalities into a fused belief assignment $\{b_{0}, b_{1}, u\}$.
In this step,
the uncertainty of one modality is employed to re-weight the belief mass of the other modality in fusion:
% the two belief assignments of the two modalities are combined to yield a more reliable joint belief assignment $M=\{b_{0}, b_{1}, u\}$:
% \begin{equation} \label{eq_5}
% M = M^{\mathbf{r}} \oplus M^{\mathbf{d}}
% \end{equation}
% where $\oplus$ denotes the Dempster’s combination \cite{DST} that uses the uncertainty of one modality to re-weight the belief mass of another modality:
% The more specific formula is as follows:
\begin{equation} \label{eq_6}
b_{k} =\frac{1}{1-C} ( b_{k}^{\mathbf{r}}\odot b_{k}^{\mathbf{d}} + u^{\mathbf{d}}\odot b_{k}^{\mathbf{r}} + u^{\mathbf{r}}\odot b_{k}^{\mathbf{d}}  ) ,
\quad u=\frac{u^{\mathbf{r}} \odot u^{\mathbf{d}}}{1-C}
\end{equation}
% where $C=\sum_{k\neq l}b_{k}^{\mathbf{r}}b_{l}^{\mathbf{d}}$ indicates the conflict between the evidences,
where $\odot$ is element-wise multiplication,
$b_k$ is the fused belief mass,
$u$ is the fused uncertainty.
$C=b_{0}^{\mathbf{r}}\odot b_{1}^{\mathbf{d}}+b_{1}^{\mathbf{r}}\odot b_{0}^{\mathbf{d}}$ measures the amount of conflict between the two assignments and
$\frac{1}{1-C}$ is the normalization term.
Observably,
\emph{the modality with a lower uncertainty has a greater impact on the fused belief mass $b_k$.}
In addition,
% the fused uncertainty $u$ becomes large \xf{iff} the uncertainties of the two modalities $u^{\mathbf{r}}, u^{\mathbf{d}}$ are both large.
the fused uncertainty $u$ becomes large iff uncertainties $u^{\mathbf{r}}, u^{\mathbf{d}}$ are both large.

% Note that the output of our network is of the size $H\times W$,
% so the multiplications in the above formula are all computed in a pixel-wise manner.

The third step aims to predict the road probability of each pixel based on the fused belief assignment.
As noted in the first step,
a belief assignment follows a Dirichlet distribution.
Thus,
we can calculate the concentration parameter $\alpha_{k}$ of Dirichlet distribution for the fused belief assignment:
% Having obtained the joint opinion $M=\{b_{0}, b_{1}, u\}$,
% the parameters $\alpha_{k}, \ k=0,1$ of the corresponding Dirichlet distribution could therefore be computed by:
\begin{equation} \label{eq_7}
S =\frac{K}{u} \quad  and  \quad  \alpha_{k} = (b_{k} \odot S) + 1
\end{equation}
where $K=2$ is the class number in road segmentation,
and $S$ is the Dirichlet strength map.
% The final output of UAF is the road probability map $P$,
% which equals to the mean of this Dirichlet distribution for the road class:
The road probability map $P$ is formulated as the mean of this Dirichlet distribution for the road class:
% where $\hat{p}_{0}$ is probability for the class non-road and $\hat{p}_{1}$ is probability for the class road, and they can be computed bu using Eq. \ref{eq_3}:
% $\hat{p}_{0} = \alpha_{0}/S$,
% $\hat{p}_{1} = \alpha_{1}/S$.
$P = \alpha_{1}/S$.
Finally,
UAF outputs the road probability map $P$ and uncertainty map $u$.

\emph{Comments}: We visualize the belief masses of road and the uncertainties obtained by our model in Fig. \ref{fig:uncertainty}.
In the first example,
UAF perceives the high uncertainty of depth modality around the distant boundary (see Fig. \ref{fig:uncertainty} (c)),
and uses low-uncertain RGB modality to eliminate the error (see Fig. \ref{fig:uncertainty} (d)).
% the depth stream generates false prediction in the distant area where the boundary is blurry as shown in Fig. \ref{fig:uncertainty} (c).
% Meanwhile, the uncertainty $u^{\mathbf{d}}$ is relative high in this area, so after fusion, the false prediction is eliminated in Fig. \ref{fig:uncertainty} (d).
In the second example,
% in Fig. \ref{fig:uncertainty} (f),
UAF gives a high uncertainty to the tree shade in RGB modality (see Fig. \ref{fig:uncertainty} (f)),
which also helps to eliminate the false prediction by using depth modality (see Fig. \ref{fig:uncertainty} (h)).
% the RGB stream generates false prediction in shadow region ,
% also a high uncertainty is generated in this area,
% and the false prediction is further eliminated as shown in Fig. \ref{fig:uncertainty} (h).

\begin{table*}[t]
\setlength{\abovecaptionskip}{-0.2cm}
\vspace{0.15cm} %调整图片与上文的垂直距离
\caption{Comparisons on the KITTI benchmark.
'MS' denotes multi-scale version.
For models using RGB and depth,
1-st results are shown in bold type,
2-nd results are shown in underline type.}
\label{table_1}
\begin{center}
\begin{tabular}{|l|c|c|c|c|c|c|c|}
\hline
Methods & Input & MaxF(\%)$\uparrow$ & PRE(\%)$\uparrow$ & REC(\%)$\uparrow$ & FPR(\%)$\downarrow$ & FNR(\%)$\downarrow$ & Runtime$\downarrow$ \\ \hline\hline
s-FCN-loc \cite{Embedding_contour}  & RGB  & 93.26   & 94.16  & 92.39  & 3.16   & 7.61   & 0.40 s      \\
MultiNet \cite{MultiNet}  & RGB   & 94.88   & 94.84  & 94.91  & 2.85   & 5.09   & 0.17 s     \\
RBNet \cite{RBNet}  & RGB  & 94.97   & 94.94  & 95.01  & 2.79   & 4.99   & 0.18 s    \\
RBANet \cite{RBANet}  & RGB  & 96.30   & 95.14  & 97.50  & 2.75   & 2.50   & 0.16 s    \\
\hline
LidCamNet \cite{LidCamNet}  & RGB + LiDAR   & 96.03   & 96.23  & 95.83  & 2.07   & 4.17   & 0.15 s    \\
CLCFNet \cite{CLCFNet}  & RGB + LiDAR  & 96.38   & 96.38  & 96.39  & 1.99   & 3.61   & 0.02 s    \\
PLARD \cite{PLARD}  & RGB + LiDAR  & 96.83   & 96.79  & 96.86  & 1.77   & 3.14   & 0.16 s    \\
PLARD (MS) \cite{PLARD}  & RGB + LiDAR  & 97.03   & 97.19  & 96.88  & 1.54   & 3.12   & 1.50 s    \\
\hline
NIM-RTFNet \cite{NIM-RTFNet}   & RGB + Depth  & 96.02   & 96.43  & 95.62  & 1.95   & 4.38   & \underline{0.05} s    \\
SNE-RoadSeg \cite{SNE-RoadSeg}  & RGB + Depth  & 96.75   & \underline{96.90}  & 96.61  & \underline{1.70}   & 3.39   & 0.10 s    \\
DFM-RTFNet \cite{DFM-RTFNet}  & RGB + Depth  & 96.78   & 96.62  & 96.93  & 1.87   & 3.07   & 0.08 s    \\
% PLARD \cite{PLARD}  & RGB + LiDAR  & 97.03   & 97.19  & 96.88  & 1.54   & 3.12   & 0.16 s    \\
SNE-RoadSeg+ \cite{SNE-RoadSeg+}  & RGB + Depth  & \textbf{97.50}   & \textbf{97.41}  & \textbf{97.58}  & \textbf{1.43}   & \textbf{2.42}   & 0.08 s    \\
% 客观性地描述
Ours  & RGB + Depth  & \underline{96.89}   & 96.51  & \underline{97.27}  & 1.94   & \underline{2.73}   & \textbf{0.02} s     \\ \hline
\end{tabular}
\end{center}
\vspace{-0.25cm}
\end{table*}

\begin{figure*}
    \centering
    \includegraphics[width=1\linewidth]{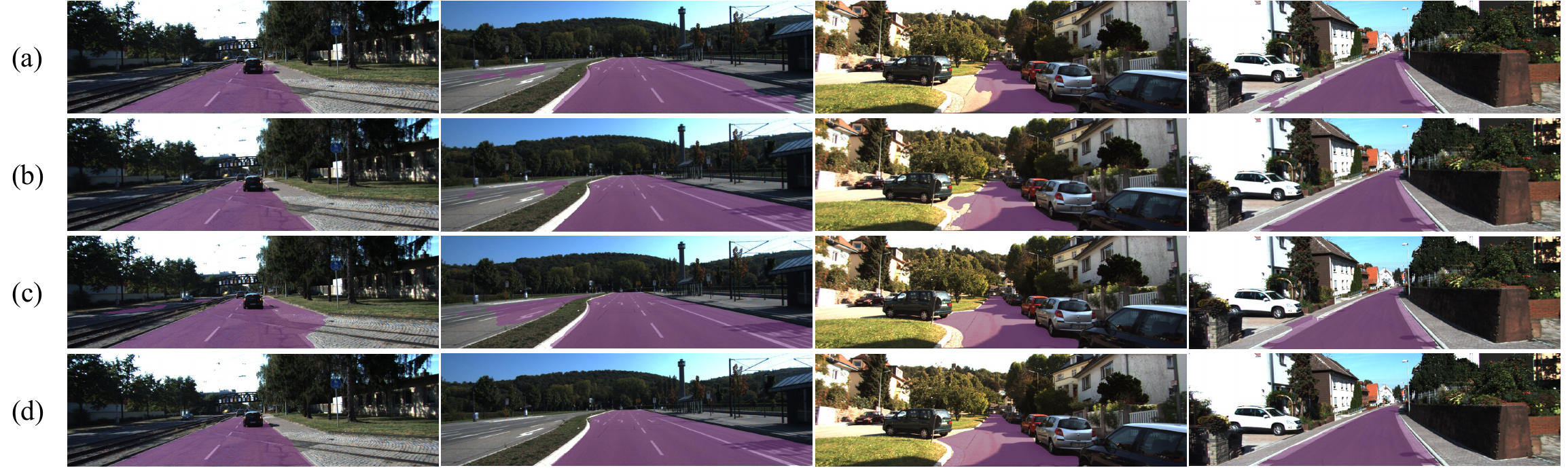}
    \vspace{-15pt}
    \caption{Example results on the KITTI benchmark:
    (a) MultiNet \cite{MultiNet},
    (b) SNE-RoadSeg \cite{SNE-RoadSeg},
    (c) PLARD \cite{PLARD}, and 
    (d) Ours.
    Note that the outputs of these methods are probability maps of road, and we use 0.5 as the threshold to generate segmentation maps.
    }
    \label{fig:KITTI_qualitative}
    \vspace{-0.25cm}
\end{figure*}

\subsection{Loss Functions for Uncertainty Learning and Fusion}
\label{sec:loss}

% Since the output of USNet is modeled as the Dirichlet distribution parameterized by $\alpha_{k}, \ k=0,1$.

% In this section,
% we give the loss function to optimize the USNet.
Given an image $I\in\mathbb{R}^{H\times W \times3}$,
we assign a one-hot label $\mathbf{y}\in\mathbb{R}^{H\times W \times2}$,
where $H,W$ denote height and width.
Here,
$\mathbf{y}_{i}\in\mathbb{R}^{2}$ is the labels of pixel $i$,
$\mathbf{y}_{i,k}\in\{0,1\}$ is a binary value with $\mathbf{y}_{i,1}$ meaning the value of pixel $i$ in the road mask,
and $\mathbf{y}_{i,0}$ for that in the non-road mask,
where $i\in\{1,2...,HW\}$.
% Let $\mathbf{y}$ denotes the label, i.e., 0 for non-road, and 1 for road,
% $\mathbf{y}_{ij,k}$ is the $k$-th class label of a pixel,
% where $i\in\{1,2...,H\}$ and $j\in\{1,2,...,W\}$.
$\bm{\alpha}=[\alpha_{0}, \alpha_{1}]$ denotes the total concentration parameter of both road and non-road,
and $\bm{\alpha}_{i}$ denotes pixel $i$ on $\bm{\alpha}$.
The Dirichlet distribution can be formed as $D(\mathbf{p}_{i}|\bm{\alpha}_{i})$,
where $\mathbf{p}_{i}$ is the class assignment probabilities of pixel $i$ on a simplex,
and $\mathbf{p}_{i,k}$ is the pixel's probability of class $k$.
Referring to \cite{Evidential_deep_learning},
the adjusted cross-entropy loss is used to guide USNet to generate more evidence for the correct prediction:
% it’s size is $H\times W$,
% and for each element $\bm{\alpha}_{ij}$ on it, we can form the Dirichlet distribution $D(\mathbf{p}_{ij}|\bm{\alpha}_{ij}), \ i=1,2...,H, \ j=1,2...,W$,
% where $\mathbf{p}_{ij}$ is the class assignment probabilities on a simplex, and $p_{ij,k}$ is the probability of $k$-th class.
% Next, we elaborate our loss function.
% Let $\mathbf{y}_{ij}$ be a one-hot encoding of ground-truth class with  $y_{ij,k}=1$ and $y_{ij,l}=0$ for all $k\neq l$, and we have the adjusted cross-entropy loss:
\begin{equation}
\begin{aligned}
L_{a}(\!\bm{\alpha}_{i}\!) \!= \!\!& \int\!\![\sum\nolimits_{k=0}^{1}\!-\mathbf{y}_{i,k}log(\!\mathbf{p}_{i,k}\!)\!]
\!\frac{1}{B(\bm{\alpha} _{i})}\!\!\prod\nolimits_{k=0}^{1}\!(\!\mathbf{p}_{i,k}\!)^{\bm{\alpha} _{i,k}\!-\!1 }d\mathbf{p}_{i} \\
 \!= \!\!& \sum\nolimits_{k=0}^{1}\mathbf{y}_{i,k}(\psi(S_{i})-\psi(\bm{\alpha}_{i,k}))
\end{aligned}
\end{equation}
where $B(\cdot)$ is the beta function and $\psi(\cdot)$ is the digamma function.
% This loss can guarantee that the generated evidence for the correct class label is higher than the evidence for other class labels.
% However, we expect the total evidence to shrink to 0 for a sample if it cannot be correctly classified,
% so the following Kullback-Leibler (KL) divergence term is introduced:
And the following Kullback-Leibler (KL) divergence \cite{Evidential_deep_learning} is used to limit the evidence for the negative label to $0$:
\begin{equation}
\begin{aligned}
KL[D(\mathbf{p}_{i}\mid \tilde{\bm{\alpha}}_{i})\parallel D( \mathbf{p}_{i}\mid \mathbf{1})]
= log (\frac{\Gamma({ \sum_{k=0}^{1}} \tilde{\alpha}_{i,k}) }{\Gamma (2){ \prod_{k=0}^{1}}\Gamma(\tilde{\alpha}_{i,k})}) 
+ \\
{\sum\nolimits_{k=0}^{1}} ( \tilde{\alpha}_{i,k}-1)[\psi( \tilde{\alpha}_{i,k}) - \psi ({ \sum\nolimits_{l=0}^{1}} \tilde{\alpha}_{i,l})]
\end{aligned}
\end{equation}
% where $\tilde{\bm{\alpha}}_{i}=\mathbf{y}_{i}+(1-\mathbf{y}_{i})\odot \bm{\alpha}_{i} $ is the Dirichlet parameters after removal of the parameter of ground-truth class for pixel $i$,
where $\tilde{\bm{\alpha}}_{i}=\mathbf{y}_{i}+(\textbf{1}-\mathbf{y}_{i})\odot \bm{\alpha}_{i} $ is a filtered Dirichlet parameter,
which is used to avoid the punishment to the positive label.
$\Gamma(\cdot)$ is the gamma function.
% Then, we get the loss that combines the adjusted cross-entropy loss and this KL term:
Then, the adjusted cross-entropy loss and this KL term are unified as follows:
\begin{equation}
L(\bm{\alpha}_{i}) = L_{a} ( \bm{\alpha}_{i}  ) + \lambda _{t}KL[ D( \mathbf{p}_{i}\mid \tilde{\bm{\alpha}}_{i}) \parallel  D( \mathbf{p}_{i}\mid \mathbf{1})]
\end{equation}
where $\lambda_{t}=min(1, t/50)\in [0, 1]$ is the balance factor,
$t$ is the index of the current training epoch.
% With this gradually increasing factor, we can avoid the output premature convergence to the uniform distribution.
% Finally,
We employ the unified loss to optimize the two subnetworks and the final prediction:
% Our network is a two-stream network, and we expect each stream can generate a reasonable opinion, so the overall loss function of our network is:
\begin{equation}
\begin{aligned}
\mathcal{L}\!=\!\!\sum_{i=1}^{HW}\!\Big(\!\beta L(\bm{\alpha}_{i})\!+\!L(\bm{\alpha}_{i}^{\mathbf{r}})\!+\!L(\bm{\alpha}_{i}^{\mathbf{d}})\!+\!\sum_{h=1}^{3}\!\!\big(L([\bm{\alpha}_{i}^{\mathbf{r}}]^h)\!+\!L([\bm{\alpha}_{i}^{\mathbf{d}}]^h)\big)\!\!\Big)
\end{aligned}
\end{equation}
where $\beta$ is a factor empirically set to $2$.
$\bm{\alpha}^{\mathbf{r}}=[e_{0}^{\mathbf{r}}+1, e_{1}^{\mathbf{r}}+1]$  and $\bm{\alpha}^{\mathbf{d}}=[e_{0}^{\mathbf{d}}+1, e_{1}^{\mathbf{d}}+1]$ denote the parameters of the Dirichlet distribution of RGB and depth subnetworks, respectively. $h$ is the index of path of the MEC module. Our network is trained end-to-end based on the unified loss.

\section{EXPERIMENTS}

In this section,
we conduct comprehensive experiments to validate the performance of the proposed network.
% The qualitative and quantitative results are conducted on two datasets.

\subsection{Experiment Setup}

\noindent
\textbf{Datasets:}
Our experiments are carried out on the KITTI road dataset \cite{KITTI} and Cityscapes dataset \cite{Cityscapes}.
The KITTI dataset is one of the most popular datasets for road segmentation.
It contains 289 training images and 290 testing images.
% Both of the two splits contain three different road scene categories including Urban Marked roads (UM), Urban Multiple Marked lanes (UMM), and Urban Unmarked roads (UU).
In the ablation study,
we split the training set into two subsets:
253 samples for training and 36 samples for validating.
The Cityscapes dataset is collected for urban scene semantic segmentation.
It contains 2975 training images and 500 validating images annotated in 19 classes,
while in our work,
we only reserve the label for the road and re-label other classes as non-road.

\noindent
\textbf{Evaluation Metrics:}
For quantitative evaluation,
we take the widely used pixel-wise segmentation metrics of road segmentation.
The metrics include maximum F1-measure (MaxF), precision (PRE), recall (REC), false-positive rate (FPR) and false-negative rate (FNR).
It is worth noting that the metrics are computed in the Birds Eye View (BEV) for the KITTI dataset as a common practice.
We also evaluate the parameters,
FLOPs, and inference time of our network.

\noindent
\textbf{Implementation Details:}
Our network is implemented using Pytorch and trained on a single NVIDIA GTX 1080Ti GPU.
A ResNet-18 \cite{ResNet} model pre-trained on ImageNet \cite{ImageNet} is employed as the backbone of USNet.
In our experiment,
the input images are resized to $384\times1248$ for the KITTI dataset and $512\times1024$ for the Cityscapes dataset.
The loss is optimized by the AdamW \cite{AdamW} optimizer.
We set the learning rate to $1e^{-4}$ for parameters of the backbone and $1e^{-3}$ for other parameters during training.
For data augmentation,
we use Gaussian blur,
Gaussian noise,
random horizontal flip,
and random color jitter on the input images.

\subsection{Evaluation Results}
In this subsection,
the results on the KITTI dataset and the Cityscapes dataset are given.

\noindent
\textbf{KITTI Benchmark:}
We report the performance on the KITTI benchmark in Table \ref{table_1}.
Our method exhibits a MaxF of $96.89\%$,
which outperforms all RGB-based methods and most RGB-LiDAR and RGB-D based methods.
The boosting is mainly owing to the use of a more efficient uncertainty-aware RGB-D fusion strategy.
Note that,
the PLARD \cite{PLARD} that has a MaxF of $97.03\%$ is trained on multiple datasets,
while our USNet is only trained on the KITTI dataset.
Although the recent proposed SNE-RoadSeg+ \cite{SNE-RoadSeg+} achieves a higher MaxF,
it is \emph{4}$\times$ slower than our method (80 ms vs. 20 ms).
Thus, our method achieves a better trade-off between the accuracy and model’s capacity.
Moreover,
the pixel-wise uncertainty given by USNet has great significance in guiding other self-driving modules, e.g., obstacle avoidance, path planning, etc.,
which is not available in existing methods.

Qualitative results are shown in Fig. \ref{fig:KITTI_qualitative}.
In detail, the first column visualizes a street where the boundary between road and sidewalk is unclear,
MultiNet \cite{MultiNet} and PLARD \cite{PLARD} suffer from the false positives,
% while USNet and SNE-RoadSeg \cite{SNE-RoadSeg} make full use of the RGB and Depth data,
% and segment the road accurately.
while USNet and SNE-RoadSeg \cite{SNE-RoadSeg} segment the road accurately with the assist of depth image.
In the second column,
other methods generate false segmentation in the left lane,
but our method avoids this mis-classification.
The third column shows a scene with over-exposure.
Our method outperforms the other three methods in this situation,
as depth data is used to compensate for the weakness of the RGB image in this scene effectively.
The last scene has the same problem as the first column,
and our method consistently generates precise segmentation.
These results prove the effectiveness and reliability of our method.

\noindent
\textbf{Cityscapes Dataset:}
We also conduct experiments on the Cityscapes dataset.
% Since the dataset is created for general scene semantic segmentation,
In the training process,
we only reserve the label for road and treat other classes as non-road.
The samples containing no road pixels are excluded from our evaluation.
% We compare our method with those algorithms which report their performance of road segmentation on this dataset.
As shown in Table \ref{table_2},
our method exceeds all other methods,
especially outperforming RBANet \cite{RBANet} by $0.27\%$ MaxF, $0.39\%$ precision, and $0.15\%$ recall,
which proves the generalization of our method.

\begin{table}[t]
\setlength{\abovecaptionskip}{-0.2cm}
\vspace{0.15cm} %调整图片与上文的垂直距离
\caption{Evaluation results on the Cityscapes validation set}
\label{table_2}
\begin{center}
\begin{tabular}{|l|c|c|c|}
\hline
Methods          & MaxF(\%) & PRE(\%) & REC(\%) \\ \hline\hline
Zohourian \emph{et al.} \cite{Superpixel} & 92.44   & 89.08  & 96.76  \\
FCN \cite{FCN}              & 94.68   & 93.69  & 95.70  \\
s-FCN-loc \cite{Embedding_contour}        & 95.36   & 94.63  & 96.11  \\
SegNet \cite{SegNet}           & 95.81   & 94.55  & 97.11  \\
RBANet \cite{RBANet}           & 98.00   & 97.87  & 98.13  \\ \hline\hline
Ours             &  \textbf{98.27}       &   \textbf{98.26}     &   \textbf{98.28}     \\ \hline
\end{tabular}
\end{center}
\vspace{-0.25cm}
\end{table}

\subsection{Ablation Studies}

\begin{table}[t]
\setlength{\abovecaptionskip}{-0.5cm}
\caption{comparison with feature fusion models}
\label{table_3}
\begin{center}
\resizebox{1\columnwidth}{!}{
\begin{tabular}{|l|c|c|c|c|c|}
\hline
Fusion settings & MaxF(\%) & PRE(\%) & REC(\%) & FPR(\%) & FNR(\%)  \\ \hline\hline
Add & 96.96   & 97.13  & 96.79  & 1.70   & 3.21   \\
Cat+Conv & 96.97   & 96.91  & 97.03  & 1.84   & 2.97   \\
RFNet \cite{RFNet} & 97.04   & 96.67  & 97.41  & 2.00   & 2.59   \\
ACNet \cite{ACNet} & 97.02   & 96.76  & 97.28  & 1.94   & 2.72   \\
SA-Gate \cite{SA-Gate} & 97.12   & 96.67  & \textbf{97.58}  & 2.00   & \textbf{2.42}   \\ \hline\hline
 Ours & \textbf{97.31}   & \textbf{97.20}  & 97.42  & \textbf{1.67}   & 2.58   \\ \hline
\end{tabular}
}
\end{center}
\vspace{-0.5cm}
\end{table}

\begin{table}[t]
\setlength{\abovecaptionskip}{-0.5cm}
\vspace{0.15cm} %调整图片与上文的垂直距离
\caption{Effectiveness analysis of proposed modules.}
\label{table_4}
\begin{center}
\resizebox{1\columnwidth}{!}{
\begin{tabular}{|l|c|c|c|c|c|}
\hline
Settings & MaxF(\%) & PRE(\%) & REC(\%) & FPR(\%) & FNR(\%)  \\ \hline\hline
 RGB   & 95.28   & 94.44  & 96.12  & 3.37   & 3.88   \\
 Depth & 96.64   & 96.14  & 97.15  & 2.32   & 2.85   \\
 RGB-D & 97.01   & 96.65  & 97.38  & 2.01   & 2.62   \\
 RGB-D+MEC & 97.16   & 96.99  & 97.33  & 1.80   & 2.67   \\
 RGB-D+UAF & 97.19   & 97.04  & 97.34  & 1.76   & 2.66   \\ \hline\hline
 Ours & \textbf{97.31} & \textbf{97.20}  & \textbf{97.42}  & \textbf{1.67}   & \textbf{2.58}   \\ \hline
\end{tabular}
}
\end{center}
\vspace{-0.25cm}
\end{table}

\begin{table}[t]
\setlength{\abovecaptionskip}{-0.2cm}
\caption{Comparison of efficiency on the KITTI dataset}
\label{table_5}
\begin{center}
\begin{tabular}{|l|c|c|c|c|}
\hline
Methods          & Params(M) & FLOPs(G) & FPS & MaxF(\%) \\ \hline\hline
MultiNet \cite{MultiNet}  & 134.3   & 406.5  & 11.6 &  94.88  \\
SNE-RoadSeg \cite{SNE-RoadSeg} & 201.3   & 1950.2  & 4.5 &  96.75 \\
PLARD \cite{PLARD}  & 76.9   & 1147.6  & 4.3 &  96.83  \\ \hline\hline
Ours  & \textbf{30.7}   & \textbf{78.2}  & \textbf{43.6} & \textbf{96.89}   \\ \hline
\end{tabular}
\end{center}
\vspace{-0.5cm}
\end{table}

% In this work, we propose a brand new network architecture for RGB-D road segmentation.
% In this section, we carried out considerable experiments to thoroughly analyze our network.
% These experiments are conducted on the KITTI dataset.
This section gives considerable experiments to thoroughly analyze our network on the KITTI dataset.

% by randomly splitting the training set into two subsets:
% 253 samples for training and 36 samples for validating.

% and as the dataset didn’t provide validation set,
% we artificially split the training set into two subsets:
% 253 samples for training and 36 samples for validating.

\noindent
\textbf{Comparison with Feature Fusion Models:}
% To verify the superiority of our method in comparison to previous works \cite{SNE-RoadSeg, NIM-RTFNet, DFM-RTFNet, SNE-RoadSeg+} that adopt a feature fusion strategy,
To verify the superiority of our fusion paradigm compared to previous works \cite{SNE-RoadSeg, NIM-RTFNet, DFM-RTFNet, SNE-RoadSeg+},
we test several feature fusion strategies by five variants. 
% In these experiments,
Each variant contains two encoders and one decoder,
and the feature maps of corresponding layers in the two encoders are fused in different ways.
% The fusion setups include ‘Add, ‘Cat+Conv’, ‘RFNet’, ‘ACNet’ and ‘SA-Gate’,
As shown in Table \ref{table_3}, 
the ‘Add’ indicates directly summing the feature of RGB and depth,
and ‘Cat+Conv’ indicates fusing by concatenation and convolution.
‘RFNet’, ‘ACNet’ and ‘SA-Gate’ indicate using the fusion strategy proposed in the semantic segmentation networks \cite{RFNet,ACNet,SA-Gate}, respectively.
% As shown in Table \ref{table_3},
Our USNet achieves the best MaxF than those models using other fusion strategies.
The comparison indicates that the different characteristics of the RGB and depth are not well perceived by directly fusing the features of the two modalities.

\noindent
\textbf{Effectiveness of Proposed Modules:}
We verify the effectiveness of each component in the proposed network,
including two subnetworks,
the multi-scale evidence collection (MEC) module,
and the uncertainty-aware fusion (UAF) module.
First,
to evaluate the necessity of the two subnetworks,
we conduct experiments based on only one modality.
As shown in Table \ref{table_4},
the RGB-based model and depth-based model achieve $95.28\%$ and $96.64\%$ in terms of MaxF,
while RGB-D based model improves the MaxF to $97.01\%$,
Note that,
in the RGB-D based model,
we fuse the segmentation result of the two subnetworks by simple addition.
Furthermore,
when MEC and UAF modules are utilized separately,
the MaxF increases by $0.15\%$ and $0.18\%$ respectively,
proving the effectiveness of MEC and UAF.
Finally,
when all these components are used,
we obtain the best MaxF of $97.31\%$.
The reason is that the MEC provides more evidence to UAF for more sufficient determination.
% This result verifies that our network could capture effective information of the RGB and Depth sources and generate robust road segmentation.

\subsection{Efficiency Analysis}
We analyze the computational efficiency of the proposed method in comparison with three open-source methods.
All speeds are gained on a same computer equipped with an NVIDIA GTX 1080Ti GPU,
and the input images are scaled into $384\times1248$ resolution for a fair comparison.
As observed in Table \ref{table_5},
the parameters and FLOPs of our network is much fewer than other methods,
which is owing to the simplified backbone and zero-parameter decoder.
Meanwhile,
our network can run at \emph{43.6} FPS,
\emph{4}$\times$ faster than MultiNet \cite{MultiNet},
and \emph{10}$\times$ faster than PLARD \cite{PLARD} and SNE-RoadSeg \cite{SNE-RoadSeg}.
In particular,
compared to SNE-RoadSeg \cite{SNE-RoadSeg},
i.e., an RGB-D based method,
USNet reduces the parameter by $85\%$, and the FLOPs by $96\%$,
but gains $0.14\%$ improvement on MaxF (see Table \ref{table_1}).
% Considering the high accuracy of our method,
Thus,
the USNet is more suitable for real-time applications intuitively
and has the potential to further boost the speed to reach the requirement of embedded platforms.

\section{Conclusion}
% 我们的方法可以和其他论文结合起来，进一步提升性能

In this work, we propose a novel low-latency RGB-D road segmentation network named USNet,
which adopts a light-weight symmetric network to separately perceive road representations based on RGB and depth data.
For collecting more valuable evidence from each subnetwork,
an MEC is proposed.
Besides, a UAF module is designed to obtain the uncertainty of the two modalities and generate the final segmentation.
All these effective designs enable our model to work satisfyingly in terms of both accuracy and computational cost.

% In this work, we propose a novel end-to-end RGB-D road segmentation network, which adopts a symmetrical two-stream network to separately make predictions based on RGB and depth modalities. Considering both RGB and depth sources contain noisy features, we utilize a multi-scale prediction module to reduce influence of noise in each modality. Further, an evidential opinion fusion module is presented to explicitly model the uncertainty of the prediction, thus we can achieve a reliable detection result. Experiments on the KITTI dataset and the Cityscapes dataset demonstrate the effectiveness of our method.

%                                   % the textheight of the last page by a suitable amount.
%                                   % This command does not take effect until the next page
%                                   % so it should come on the page before the last. Make
%                                   % sure that you do not shorten the textheight too much.

% \section{Acknowledgments}
% This work was supported by the national key R \& D program intergovernmental international science and technology innovation cooperation project  2021YFE0101600.

\balance
\bibliography{ref}

\end{document}